%% file: RJwrapper.tex
\documentclass[a4paper]{report}
\usepackage[utf8]{inputenc}
\usepackage[T1]{fontenc}
\usepackage{RJournal}
\usepackage{amsmath,amssymb,array}
\usepackage{booktabs}
  \usepackage{amsmath}
  \usepackage{cleveref}
  \usepackage{float}
  \usepackage{xspace}

\providecommand{\tightlist}{%
  \setlength{\itemsep}{0pt}\setlength{\parskip}{0pt}}

\begin{document}

%% do not edit, for illustration only
\sectionhead{}
\volume{XX}
\volnumber{YY}
\year{20ZZ}
\month{AAAA}

\begin{article}
  \input{proellochs-feuerriegel}

\end{article}

\end{document}

%% file: proellochs-feuerriegel.tex
% !TeX root = RJwrapper.tex
\title{Reinforcement Learning in R}
\author{by Nicolas Pröllochs, Stefan Feuerriegel}

\maketitle

\abstract{%
Reinforcement learning refers to a group of methods from artificial
intelligence where an agent performs learning through trial and error.
It differs from supervised learning, since reinforcement learning
requires no explicit labels; instead, the agent interacts continuously
with its environment. That is, the agent starts in a specific state and
then performs an action, based on which it transitions to a new state
and, depending on the outcome, receives a reward. Different strategies
(e.g.~Q-learning) have been proposed to maximize the overall reward,
resulting in a so-called policy, which defines the best possible action
in each state. Mathematically, this process can be formalized by a
Markov decision process and it has been implemented by packages in R;
however, there is currently no package available for reinforcement
learning. As a remedy, this paper demonstrates how to perform
reinforcement learning in R and, for this purpose, introduces the
\pkg{ReinforcementLearning} package. The package provides a remarkably
flexible framework and is easily applied to a wide range of different
problems. We demonstrate its use by drawing upon common examples from
the literature (e.g.~finding optimal game strategies).
}

\newcommand{\TODO}[1]{{\color{red}#1}}
\newcommand\ie{i.\,e.\xspace}
\newcommand\eg{e.\,g.\xspace}

\hypertarget{introduction}{%
\section{Introduction}\label{introduction}}

Reinforcement learning represents an approach to solving sequential and
usually stochastic decision-making problems \citep{Sutton.1998}. In
contrast to supervised machine learning, it requires no explicit labels
indicating what the correct solution is; instead, it interacts with the
environment and learns through trial and error. In many cases, this
approach appears quite natural by mimicking the fundamental way humans
learn and can thereby infer a rewarding strategy on its own. For this
purpose, reinforcement learning assumes an agent that sequentially
undertakes different actions based on which it transitions between
states. Here it receives only limited feedback in the form of a
numerical reward that is to be maximized over time. The agent usually
performs the action promising the highest long-term reward, but will
sometimes follow a random choice as part of its exploration.
Mathematically, the underlying problem relies upon the Markov decision
process \citep[MDP in short; cf.][]{Bellman.1957}, which we formalize in
Section~\ref{problem-specification}. To this end, we refer to a detailed
introduction to the literature, such as \citet{Sutton.1998}.

Reinforcement learning can be applied to a wide range of sequential
decision-making problems from numerous domains. We detail a few
illustrative examples in the following. Among others, it is particularly
popular in robotics \citep[e.g.][]{Smart.2002,Mataric.1997} where
reinforcement learning navigates robots with the aim of avoiding
collisions with obstacles \citep{Beom.1995}. Analogously, reinforcement
learning facilitates decision-making problems underlying control theory,
such as improving the performance of elevators \citep{Barto.1996}.
Beyond engineering settings, its methodology also applies to the field
of game theory \citep{Erev.1998,Erev.1998b} and, similarly, allows
algorithms to derive strategies for computer games \citep{Mnih.2015}.
Beyond that, it also helps in optimizing financial trading
\citep{Nevmyvaka.2006} or tuning hyperparameters in machine learning
algorithms \citep{Zoph.2016}. In the context of computational
linguistics, \citet{Scheffler.2002} utilize reinforcement learning to
enhance human-computer dialogs, while \citet{DSS.2016} derive policies
for detecting negation scopes in order to improve the accuracy of
sentiment analysis. Altogether, reinforcement learning helps to address
all kind of problems involving sequential decision-making.

Reinforcement learning techniques can primarily be categorized in two
groups, namely, model-based and model-free approaches
\citep{Sutton.1998}. The former, \emph{model-based} algorithms, rely on
explicit models of the environment that fully describe the probabilities
of state transitions, as well as the consequences of actions and the
associated rewards. Specifically, corresponding algorithms are built
upon explicit representations of the environment given in the form of
Markov decision processes. These MDPs can be solved by various,
well-known algorithms, including value iteration and policy iteration,
in order to derive the optimal behavior of an agent. These algorithms
are also implemented within the statistical software R. For instance,
the package \pkg{MDPtoolbox} solves such models based on an explicit
formalization of the MDP, i.e.~settings in which the transition
probabilities and reward functions are known a priori
\citep{MDPtoolbox.2017}.

The second category of reinforcement learning comprises
\emph{model-free} approaches that forgo any explicit knowledge regarding
the dynamics of the environment. These approaches learn the optimal
behavior through trial-and-error by directly interacting with the
environment. In this context, the learner has no explicit knowledge of
either the reward function or the state transition function
\citep{Hu.2003}. Instead, the optimal behavior is iteratively inferred
from the consequences of actions on the basis of past experience. As a
main advantage, this method is applicable to situations in which the
dynamics of the environment are unknown or too complex to evaluate.
However, the available tools in R are not yet living up to the needs of
users in such cases. In fact, there is currently no package available
that allows one to perform model-free reinforcement learning in R.
Hence, users that aim to teach optimal behavior through trial-and-error
learning must implement corresponding learning algorithms in a manual
way.

As a remedy, we introduce the \pkg{ReinforcementLearning} package for R,
which allows an agent to learn the optimal behavior based on sampling
experience consisting of states, actions and rewards
\citep{ReinforcementLearning.2017}\footnote{The \pkg{ReinforcementLearning} R package is available from the Comprehensive R Archive Network (CRAN) at \url{http://cran.r-project.org/package=ReinforcementLearning}.}.
The training examples for reinforcement learning can originate from
real-world applications, such as sensor data. In addition, the package
is shipped with the built-in capability to sample experience from a
function that defines the dynamics of the environment. In both cases,
the result of the learning process is a highly interpretable
reinforcement learning policy that defines the best possible action in
each state. The package provides a remarkably flexible framework, which
makes it readily applicable to a wide range of different problems. Among
other functions, it allows one to customize a variety of learning
parameters and elaborates on how to mitigate common performance in
common solution algorithms (e.g.~experience replay).

As our main contribution, we present a novel package for implementing
reinforcement learning in R. This can have a considerable impact for
practitioners in the field of artificial intelligence and, hence,
immediately reveals manifold implications: one can utilize our package
to implement self-learning systems. We thus detail the applicability and
usage of the package throughout this paper. For this purpose, we
demonstrate the package by drawing upon common examples from the
literature.

The remainder of this paper is structured as follows.
Section~\ref{background} specifies a mathematical formalization of
reinforcement learning and details how it solves sequential
decision-making problems in a dynamic environment. Subsequently, we
introduce the \pkg{ReinforcementLearning} package and demonstrate its
main functionality and routines (Section~\ref{package-features}).
Section~\ref{illustrative-example} provides a illustrative example in
which a robot needs to find the optimal movements in a maze, while
Section~\ref{conclusion} concludes.

\hypertarget{background}{%
\section{Background}\label{background}}

\hypertarget{markov-decision-process}{%
\subsection{Markov decision process}\label{markov-decision-process}}

A Markov decision process \citep{Bellman.1957} is defined by a 5-tuple
\((S, A, P, R, \gamma)\) where \(S\) is a set of states. Let \(s \in S\)
refer to the current state. Then, in each state, one can undertake an
action \(a \in A\) from the set of actions \(A\). Here one sometimes
assumes only a subset \(A_s \subset A\) dependent on a state \(s\). By
taking an action, the state transitions from \(s\) to \(s'\) with a
probability \(P_a(s, s')\). Upon arriving in a new state \(s'\) due to
action \(a\), one receives a reward given by \(R_a(s, s')\). Lastly, the
parameter \(\gamma \in [0, 1]\) denotes a discount factor, which
controls how important the current reward is in future settings.

As a result, we observe a few characteristics of our problem: first, the
problem is time-dependent in the sense that choosing an action
influences our future state. Second, the formulation implies a
stochastic setting where the new state is not deterministic but can only
be predicted with a certain probability. Third, a model-free setup
requires that parameters \(P\) and \(R\) are not known a priori.
Instead, once can merely learn the state transitions and expected
rewards from interacting with the environment. This motivates the
following approach underlying reinforcement learning.

\hypertarget{reinforcement-learning}{%
\subsection{Reinforcement learning}\label{reinforcement-learning}}

\hypertarget{problem-specification}{%
\subsubsection{Problem specification}\label{problem-specification}}

\begin{figure}[h]
\centering
    \includegraphics[width=.8\linewidth]{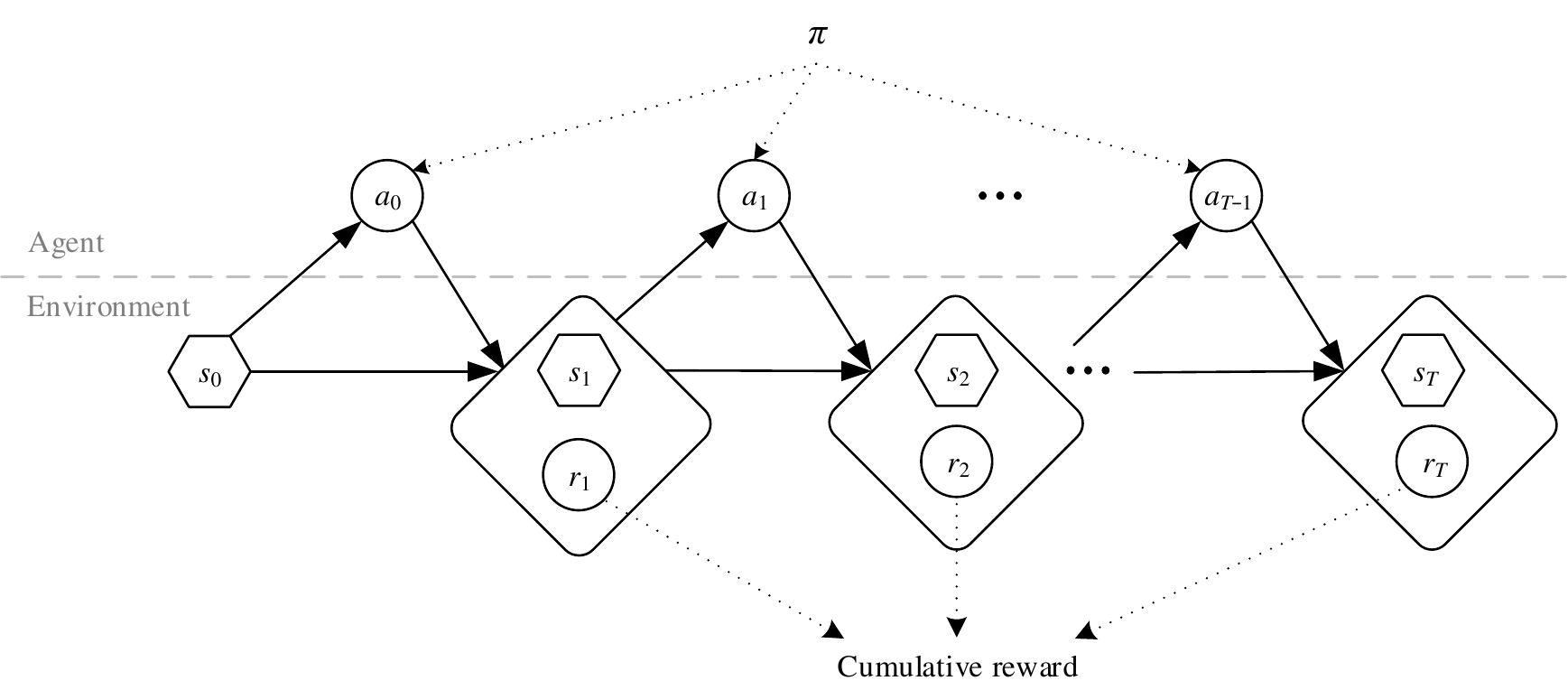}
\caption{The agent-environment interaction in reinforcement learning.}
\label{fig:agent_environment}
\end{figure}

Reinforcement learning assumes a so-called \emph{agent} which interacts
with the \emph{environment} over a sequence of discrete steps. Thereby,
the agent develops an understanding of the environment (mathematically,
\(P\) and \(R\)), while seeking to maximize the cumulative reward over
time. Formally, we rewrite the problem formulation as an iterative
process, as visualized in \Cref{fig:agent_environment}. Based on the
current state \(s_i\) in step \(i\), the agent picks an action
\(a_i \in A_{s_i} \subseteq A\) out of the set of available actions for
the given state \(s_i\). Subsequently, the agent receives feedback
related to its decision in the form of a numerical reward \(r_{i+1}\),
after which it then moves to the next state \(s_{i+1}\).

The resulting objective of reinforcement learning is then to determine a
\emph{policy} \(\pi(s_i, a_i)\) that defines which action \(a_i\) to
take in state \(s_i\). In other words, the goal of the agent is to learn
the optimal policy function for all states and actions. To do so, the
approach maintains a \emph{state-action function} \(Q(s_i, a_i)\),
specifying the expected reward for each possible action \(a_i\) in state
\(s_i\). This knowledge can then be used to infer the optimal behavior,
i.e.~the policy that maximizes the expected reward from any state. The
\emph{optimal policy} \(\pi^*\) maximizes the overall expected reward
or, more specifically, the sum of discounted rewards given by
\begin{equation}
\pi^\ast = \max_{\pi}\;E \left[ \sum_{i=0}^{\infty} \gamma^i r_t \,\|\, s_0 = s \right] . 
\end{equation} In practice, one chooses the actions \(a_i\) that
maximize \(Q(s_i,a_i)\). A common approach to computing this policy is
via \emph{Q-learning} \citep{Watkins.1992}. As opposed to a model-based
approach, \(Q\)-learning has no explicit knowledge of either the reward
function or the state transition function \citep{Hu.2003}. Instead, it
iteratively updates its action-value \(Q(s_i,a_i)\) based on past
experience. Alternatives are, for instance, SARSA or temporal-difference
(TD) learning, but these are less common in practice nowadays.

\hypertarget{exploration}{%
\subsubsection{Exploration}\label{exploration}}

The above descriptions specify how to choose the optimal action after
having computed \(Q\). For this purpose, reinforcement learning performs
an exploration in which it randomly selects action without reference to
\(Q\). Thereby, it collects knowledge about how rewarding actions are in
each state. One such exploration method is \(\varepsilon\)-greedy, where
the agent chooses a random action uniformly with probability
\(\varepsilon \in (0, 1)\) and otherwise follows the one with the
highest expected long-term reward. Mathematically, this yields
\begin{equation}
a_i = \begin{cases}
\max_{a \in A_{s_i}}\;Q(s_i, a_i), & \text{with probability } 1- \varepsilon,\\
\text{pick random } a \in A_{s_i},  & \text{with probability } \varepsilon .
\end{cases}
\end{equation} For a description of alternative heuristics, we refer to
\citet{Sutton.1998}.

\hypertarget{q-learning}{%
\subsubsection{Q-learning}\label{q-learning}}

The Q-learning algorithm \citep{Watkins.1992} starts with arbitrary
initialization of \(Q(s_i,a_i)\) for all \(s_i \in S\) and
\(a_i \in A\). The agent then proceeds with the aforementioned
interaction steps: (1)~it observes the current state \(s_i\), (2)~it
performs an action \(a_{i}\), (3)~it receives a subsequent reward
\(r_{i+1}\) after which it observes the next state \(s_{i+1}\). Based on
this information, it then estimates the optimal future value
\(Q(s_{i+1}, a_{i+1})\), i.e.~the maximum possible reward across all
actions available in \(s_{i+1}\). Subsequently, it uses this knowledge
to update the value of \(Q(s_i,a_i)\). This is accomplished via the
update rule \begin{equation}
%\begin{split}
Q(s_i,a_i) \gets \, Q(s_i,a_i) + \alpha\left[r_{i+1}  + \gamma \max_{a_{i+1} \in A_{s_i}} Q(s_{i+1}, a_{i+1}) - Q(s_i,a_i)\right],
%\end{split}
\end{equation} with a given learning rate \(\alpha\) and discount factor
\(\gamma\). Thus, Q-learning learns an optimal policy while following
another policy for action selection. This feature is called
\emph{off-policy} learning. More detailed mathematical explanations and
a thorough survey of this and other reinforcement learning methods are
given in, for example, \citet{Kaelbling.1996} and \citet{Sutton.1998}.

\hypertarget{extensions}{%
\subsection{Extensions}\label{extensions}}

\hypertarget{batch-learning}{%
\subsubsection{Batch learning}\label{batch-learning}}

Reinforcement learning generally maximizes the sum of expected rewards
in an agent-environment loop. However, this setup often needs many
interactions until convergence to an optimal policy. As a result, the
approach is hardly feasible in complex, real-world scenarios. A suitable
remedy comes in the form of so-called
\emph{batch reinforcement learning} \citep{Lange.2012}, which
drastically reduces the computation time in most settings. Different
from the previous setup, the agent no longer interacts with the
environment by processing single state-action pairs for which it updates
the \(Q\)-table on an individual basis. Instead, it directly receives a
set of tuples \((s_i,a_i,r_{i+1},s_{i+1})\) sampled from the
environment. After processing this batch, it updates the \(Q\)-table, as
well as the policy. This procedure commonly results in greater
efficiency and stability of the learning process, since the noise
attached to individual explorations cancels out as one combines several
explorations in a batch.

\begin{figure}[H]
\centering
    \includegraphics[width=\linewidth]{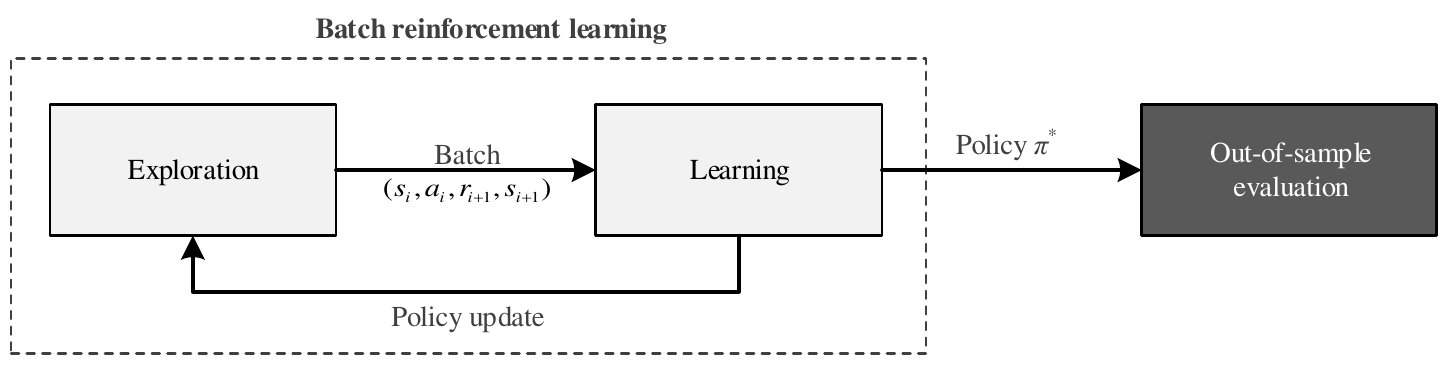}
\caption{Batch reinforcement learning.}
\label{fig:rl_batch}
\end{figure}

The batch process is grouped into three phases (see
\Cref{fig:rl_batch}). In the first phase, the learner explores the
environment by collecting transitions with an arbitrary sampling
strategy, e.g.~using a purely random policy or \(\varepsilon\)-greedy.
These data points can be, for example, collected from a running system
without the need for direct interaction. In a second step, the stored
training examples then allow the agent to learn a state-action function,
e.g.~via Q-learning. The result of the learning step is the best
possible policy for every state transition in the input data. In a final
step, the learned policy can be applied to the system. Here the policy
remains fixed and is no longer
improved.\footnote{Alternatively, the policy can be used for validation purposes or to collect new data points (e.g. in order to iteratively improve the current policy). This variant of reinforcement learning is referred to as \emph{growing batch reinforcement learning}. In this method, the agent alternates between exploration and training phases, which is oftentimes the model of choice when applying reinforcement learning to real-world problems. It is also worth noting that this approach is similar to the general reinforcement learning problem: the agent improves its policy \emph{while} interacting with the system.}

\hypertarget{experience-replay}{%
\subsubsection{Experience replay}\label{experience-replay}}

Another performance issue that can result in poor performance in
reinforcement learning cases is that information might not easily spread
through the whole state space. For example, an update of Q-value in step
\(i\) of the state-action pair \((s_i, a_i)\) might update the values of
\((s_{i-1},a)\) for all \(a \in A_{s_i}]\). However, this update is not
back-propagated immediately to all the involved preceding states.
Instead, preceding states are only updated the next time they are
visited. These \emph{local} updates can result in serious performance
issues since additional interactions are needed to spread
already-available information through the whole state space.

As a remedy, the concept of \emph{experience replay} allows
reinforcement learning agents to remember and reuse experiences from the
past \citep{Lin.1992}. The underlying idea is to speed up convergence by
replaying observed state transitions repeatedly to the agent, as if they
were new observations collected while interacting with a system.
Experience replay thus makes more efficient use of this information by
resampling the connection between individual states. As its main
advantage, experience replay can speed up convergence by allowing for
the back-propagation of information from updated states to preceding
states without further interaction.

\hypertarget{package-features}{%
\section{Package features}\label{package-features}}

Even though reinforcement learning has become prominent in machine
learning, the R landscape is not living up to the needs of researchers
and practitioners. The \pkg{ReinforcementLearning} package is intended
to close this gap and offers the ability to perform model-free
reinforcement learning in a highly customizable framework. The following
sections present its usage and main functionality.

Before running all subsequent code snippets, one first needs to load the
package via:

\begin{Schunk}
\begin{Sinput}
library("ReinforcementLearning")
\end{Sinput}
\end{Schunk}

\hypertarget{data-preparation}{%
\subsection{Data preparation}\label{data-preparation}}

The \pkg{ReinforcementLearning} package utilizes different mechanisms
for reinforcement learning, including Q-learning and experience replay.
It thereby learns an optimal policy based on past experience in the form
of sample sequences consisting of states, actions and rewards.
Consequently, each training example consists of a state-transition tuple
\((s_i, a_i, r_{i+1}, s_{i+1})\) as follows:

\begin{itemize}
\tightlist
\item
  \(s_i\) is the current environment state.
\item
  \(a_i\) denotes the selected action in the current state.
\item
  \(r_{i+1}\) specifies the immediate reward received after
  transitioning from the current state to the next state.
\item
  \(s_{i+1}\) refers to the next environment state.
\end{itemize}

The training examples for reinforcement learning can (1)~be collected
from an external source and inserted into a tabular data structure, or
(2)~generated dynamically by querying a function that defines the
behavior of the environment. In both cases, the corresponding input must
follow the same tuple structure \((s_i, a_i, r_{i+1}, s_{i+1})\). We
detail both variants in the following.

\hypertarget{learning-from-pre-defined-observations}{%
\subsubsection{Learning from pre-defined
observations}\label{learning-from-pre-defined-observations}}

This approach is beneficial when the input data is pre-determined or one
wants to train an agent that replicates past behavior. In this case, one
merely needs to insert a tabular data structure with past observations
into the package. This might be the case when the state-transition
tuples have been collected from an external source, such as sensor data,
and one wants to learn an agent by eliminating further interaction with
the environment.

\textbf{Example.} The following example shows the first five
observations of a representative dataset containing game states of
randomly sampled tic-tac-toe
games\footnote{The tic-tac-toe dataset contains game states of 406,541 randomly-sampled tic-tac-toe games and is included in the \pkg{ReinforcementLearning} R package. All states are observed from the perspective of player X who is also assumed to have played first. The player who succeeds in placing three of their marks in a horizontal, vertical, or diagonal row wins the game. Reward for player X is +1 for 'win', 0 for 'draw', and -1 for 'loss'.}.
In this dataset, the first column contains a representation of the
current board state in a match. The second column denotes the observed
action of player X in this state, whereas the third column contains a
representation of the resulting board state after performing the action.
The fourth column specifies the resulting reward for player X. This
dataset is thus sufficient as input for learning the agent.

\begin{Schunk}
\begin{Sinput}
data("tictactoe")
head(tictactoe, 5)
\end{Sinput}
\begin{Soutput}
#>       State Action NextState Reward
#> 1 .........     c7 ......X.B      0
#> 2 ......X.B     c6 ...B.XX.B      0
#> 3 ...B.XX.B     c2 .XBB.XX.B      0
#> 4 .XBB.XX.B     c8 .XBBBXXXB      0
#> 5 .XBBBXXXB     c1 XXBBBXXXB      0
\end{Soutput}
\end{Schunk}

\hypertarget{dynamic-learning-from-an-interactive-environment-function}{%
\subsubsection{Dynamic learning from an interactive environment
function}\label{dynamic-learning-from-an-interactive-environment-function}}

An alternative strategy is to define a function that mimics the behavior
of the environment. One can then learn an agent that samples experience
from this function. Here the environment function takes a state-action
pair as input. It then returns a list containing the name of the next
state and the reward. In this case, one can also utilize R to access
external data sources, such as sensors, and execute actions via common
interfaces. The structure of such a function is represented by the
following pseudocode:

\begin{verbatim}
environment <- function(state, action) {
  ...
  return(list("NextState" = newState,
              "Reward" = reward))
}
\end{verbatim}

After specifying the environment function, we can use
\code{sampleExperience()} to collect random sequences from it. Thereby,
the input specifies number of samples (\(N\)), the environment function,
the set of states (i.e. \(S\)) and the set of actions (i.e. \(A\)). The
return value is then a data frame containing the experienced state
transition tuples \((s_i, a_i, r_{i+1}, s_{i+1})\) for
\(i = 1, \ldots, N\). The following code snippet shows how to generate
experience from an exemplary environment
function.\footnote{The exemplary grid world environment function is included in the package and can be loaded via \code{gridworldEnvironment}. We will detail this example later in Section~\ref{illustrative-example}.}

\begin{Schunk}
\begin{Sinput}
# Define state and action sets
states <- c("s1", "s2", "s3", "s4")
actions <- c("up", "down", "left", "right")

env <- gridworldEnvironment

# Sample N = 1000 random sequences from the environment
data <- sampleExperience(N = 1000, 
                         env = env, 
                         states = states, 
                         actions = actions)
\end{Sinput}
\end{Schunk}

\hypertarget{learning-phase}{%
\subsection{Learning phase}\label{learning-phase}}

\hypertarget{general-setup}{%
\subsubsection{General setup}\label{general-setup}}

The routine \code{ReinforcementLearning()} bundles the main
functionality, which teaches a reinforcement learning agent using the
previous input data. For this purpose, it requires the following
arguments: (1)~A \emph{data} argument that must be a data frame object
in which each row represents a state transition tuple
\((s_i, a_i, r_{i+1}, s_{i+1})\). (2)~The user is required to specify
the column names of the individual tuple elements within \emph{data}.

The following pseudocode demonstrates the usage for pre-defined data
from an external source, while Section~\ref{illustrative-example}
details the interactive setup. Here the parameters \emph{s}, \emph{a},
\emph{r} and \emph{s\_new} contain strings specifying the corresponding
column names in the data frame \emph{data}.

\begin{Schunk}
\begin{Sinput}
# Load dataset
data("tictactoe")

# Perform reinforcement learning
model <- ReinforcementLearning(data = tictactoe, 
                               s = "State", 
                               a = "Action", 
                               r = "Reward", 
                               s_new = "NextState", 
                               iter = 1)
\end{Sinput}
\end{Schunk}

\hypertarget{parameter-configuration}{%
\subsubsection{Parameter configuration}\label{parameter-configuration}}

Several parameters can be provided to \code{ReinforcementLearning()} in
order to customize the learning behavior of the agent.

\begin{itemize}
\tightlist
\item
  \textbf{alpha} The learning rate, set between 0 and 1. Setting it to 0
  means that the Q-values are never updated and, hence, nothing is
  learned. Setting a high value, such as 0.9, means that learning can
  occur quickly.
\item
  \textbf{gamma} Discount factor, set between 0 and 1. Determines the
  importance of future rewards. A factor of 0 will render the agent
  short-sighted by only considering current rewards, while a factor
  approaching 1 will cause it to strive for a greater reward over the
  long run.
\item
  \textbf{epsilon} Exploration parameter, set between 0 and 1. Defines
  the exploration mechanism in \(\varepsilon\)-greedy action selection.
  In this strategy, the agent explores the environment by selecting an
  action at random with probability \(\varepsilon\). Alternatively, the
  agent exploits its current knowledge by choosing the optimal action
  with probability \(1-\varepsilon\). This parameter is only required
  for sampling new experience based on an existing policy.
\item
  \textbf{iter} Number of repeated learning iterations. This must be an
  integer greater than 0. The default is set to 1. This parameter is
  passed directly to \code{ReinforcementLearning()}.
\end{itemize}

The learning parameters \textbf{alpha}, \textbf{gamma}, and
\textbf{epsilon} must be provided in an optional \textbf{control} object
passed to the \code{ReinforcementLearning()} function.

\begin{Schunk}
\begin{Sinput}
# Define control object
control <- list(alpha = 0.1, gamma = 0.1, epsilon = 0.1)

# Pass learning parameters to reinforcement learning function
model <- ReinforcementLearning(data, iter = 10, control = control)
\end{Sinput}
\end{Schunk}

\hypertarget{diagnostics}{%
\subsection{Diagnostics}\label{diagnostics}}

The result of the learning process is an object of type \code{"rl"} that
contains the state-action table and an optimal policy with the best
possible action in each state. The command \code{policy(model)} shows
the optimal policy, while \code{print(model)} outputs the state-action
table, i.e.~the Q-value of each state-action pair. In addition,
\code{summary(model)} prints further model details and summary
statistics.

\begin{Schunk}
\begin{Sinput}
# Print policy
policy(model)

# Print state-action table
print(model)

# Print summary statistics
summary(model)
\end{Sinput}
\end{Schunk}

\hypertarget{illustrative-example}{%
\section{Illustrative example}\label{illustrative-example}}

This section demonstrates the capabilities of the
\pkg{ReinforcementLearning} package with the help of a practical
example.

\hypertarget{problem-definition}{%
\subsection{Problem definition}\label{problem-definition}}

\begin{figure}[H]
\centering
    \includegraphics[width=.2\linewidth]{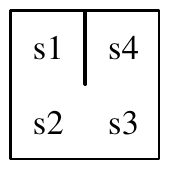}
\caption{Simulated $2\times 2$ grid.}
\label{fig:grid_2x2}
\end{figure}

Our practical example aims at teaching optimal movements to a robot in a
grid-shaped maze \citep{Sutton.1998}. Here the agent must navigate from
a random starting position to a final position on a simulated
\(2 \times 2\) grid (see \Cref{fig:grid_2x2}). Each cell on the grid
reflects one state, yielding a total of 4 different states. In each
state, the agent can perform one out of four possible actions, i.e.~to
move up, down, left, or right, with the only restriction being that it
must remain on the grid. In other words, the grid is surrounded by a
wall, which makes it impossible for the agent to move off the grid. A
wall between \emph{s1} and \emph{s4} hinders direct movements between
these states. Finally, the reward structures is as follows: each
movement leads to a negative reward of -1 in order to penalize routes
that are not the shortest path. If the agent reaches the goal position,
it earns a reward of 10.

\hypertarget{defining-an-environment-function}{%
\subsection{Defining an environment
function}\label{defining-an-environment-function}}

We first define the sets of available states (\emph{states}) and actions
(\emph{actions}).

\begin{Schunk}
\begin{Sinput}
# Define state and action sets
states <- c("s1", "s2", "s3", "s4")
actions <- c("up", "down", "left", "right")
\end{Sinput}
\end{Schunk}

We then rewrite the above problem formulation into the following
environment function. As previously mentioned, this function must take a
state and an action as input. The if-conditions determine the current
combination of state and action. In our example, the state refers to the
agent's position on the grid and the action denotes the intended
movement. Based on these, the function decides upon the next state and a
numeric reward. These together are returned as a list.

\begin{Schunk}
\begin{Sinput}
# Load built-in environment function for 2x2 gridworld 
env <- gridworldEnvironment
print(env)
\end{Sinput}
\begin{Soutput}
#> function (state, action) 
#> {
#>     next_state <- state
#>     if (state == state("s1") && action == "down") 
#>         next_state <- state("s2")
#>     if (state == state("s2") && action == "up") 
#>         next_state <- state("s1")
#>     if (state == state("s2") && action == "right") 
#>         next_state <- state("s3")
#>     if (state == state("s3") && action == "left") 
#>         next_state <- state("s2")
#>     if (state == state("s3") && action == "up") 
#>         next_state <- state("s4")
#>     if (next_state == state("s4") && state != state("s4")) {
#>         reward <- 10
#>     }
#>     else {
#>         reward <- -1
#>     }
#>     out <- list(NextState = next_state, Reward = reward)
#>     return(out)
#> }
#> <bytecode: 0x000000001654bcc0>
#> <environment: namespace:ReinforcementLearning>
\end{Soutput}
\end{Schunk}

\hypertarget{learning-an-optimal-policy}{%
\subsection{Learning an optimal
policy}\label{learning-an-optimal-policy}}

After having specified the environment function, we can use the built-in
\code{sampleExperience()} function to sample observation sequences from
the environment. The following code snippet generates a data frame
\emph{data} containing 1000 random state-transition tuples
\((s_i, a_i, r_{i+1}, s_{i+1})\).

\begin{Schunk}
\begin{Sinput}
# Sample N = 1000 random sequences from the environment
data <- sampleExperience(N = 1000, 
                         env = env, 
                         states = states, 
                         actions = actions)
head(data)
\end{Sinput}
\begin{Soutput}
#>   State Action Reward NextState
#> 1    s1   down     -1        s2
#> 2    s1   down     -1        s2
#> 3    s3   down     -1        s3
#> 4    s1     up     -1        s1
#> 5    s3  right     -1        s3
#> 6    s4  right     -1        s4
\end{Soutput}
\end{Schunk}

We can now use the observation sequence in \emph{data} in order to learn
the optimal behavior of the agent. For this purpose, we first customize
the learning behavior of the agent by defining a control object. We
follow the default parameter choices and set the learning rate
\emph{alpha} to 0.1, the discount factor \emph{gamma} to 0.5 and the
exploration greediness \emph{epsilon} to 0.1. Subsequently, we use the
\code{ReinforcementLearning()} function to learn the best possible
policy for the the input data.

\begin{Schunk}
\begin{Sinput}
# Define reinforcement learning parameters
control <- list(alpha = 0.1, gamma = 0.5, epsilon = 0.1)

# Perform reinforcement learning
model <- ReinforcementLearning(data, 
                               s = "State", 
                               a = "Action", 
                               r = "Reward", 
                               s_new = "NextState", 
                               control = control)
\end{Sinput}
\end{Schunk}

\hypertarget{evaluating-policy-learning}{%
\subsection{Evaluating policy
learning}\label{evaluating-policy-learning}}

The \code{ReinforcementLearning()} function returns an \code{"rl"}
object. We can evoke \code{policy(model)} in order to display the policy
that defines the best possible action in each state. Alternatively, we
can use \code{print(model)} in order to write the entire state-action
table to the screen, i.e.~the Q-value of each state-action pair.
Evidently, the agent has learned the optimal policy that allows it to
take the shortest path from an arbitrary starting position to the goal
position \emph{s4}.

\begin{Schunk}
\begin{Sinput}
# Print policy
policy(model)
\end{Sinput}
\begin{Soutput}
#>      s1      s2      s3      s4 
#>  "down" "right"    "up"  "left"
\end{Soutput}
\begin{Sinput}
# Print state-action function
print(model)
\end{Sinput}
\begin{Soutput}
#> State-Action function Q
#>         right         up       down       left
#> s1 -0.7210267 -0.7275138  0.6573701 -0.7535771
#> s2  3.5286336 -0.7862925  0.6358511  0.6607884
#> s3  3.5460468  9.1030684  3.5353220  0.6484856
#> s4 -1.8697756 -1.8759779 -1.8935405 -1.8592323
#> 
#> Policy
#>      s1      s2      s3      s4 
#>  "down" "right"    "up"  "left" 
#> 
#> Reward (last iteration)
#> [1] -340
\end{Soutput}
\end{Schunk}

Ultimately, we can use \code{summary(model)} to inspect the model
further. This command outputs additional diagnostics regarding the model
such as the number of states and actions. Moreover, it allows us to
analyze the distribution of rewards. For instance, we see that the total
reward in our sample (i.e.~the sum of the rewards column \(r\)) is
highly negative. This indicates that the random policy used to generate
the state transition samples deviates from the optimal case. Hence, the
next section explains how to apply and update a learned policy with new
data samples.

\begin{Schunk}
\begin{Sinput}
# Print summary statistics
summary(model)
\end{Sinput}
\begin{Soutput}
#> 
#> Model details
#> Learning rule:           experienceReplay
#> Learning iterations:     1
#> Number of states:        4
#> Number of actions:       4
#> Total Reward:            -340
#> 
#> Reward details (per iteration)
#> Min:                     -340
#> Max:                     -340
#> Average:                 -340
#> Median:                  -340
#> Standard deviation:      NA
\end{Soutput}
\end{Schunk}

\hypertarget{applying-a-policy-to-unseen-data}{%
\subsection{Applying a policy to unseen
data}\label{applying-a-policy-to-unseen-data}}

We now apply an existing policy to unseen data in order to evaluate the
out-of-sample performance of the agent. The following example
demonstrates how to sample new data points from an existing policy. The
result yields a column with the best possible action for each given
state.

\begin{Schunk}
\begin{Sinput}
# Example data
data_unseen <- data.frame(State = c("s1", "s2", "s1"), 
                          stringsAsFactors = FALSE)

# Pick optimal action
data_unseen$OptimalAction <- predict(model, data_unseen$State)

data_unseen
\end{Sinput}
\begin{Soutput}
#>   State OptimalAction
#> 1    s1          down
#> 2    s2         right
#> 3    s1          down
\end{Soutput}
\end{Schunk}

\hypertarget{updating-an-existing-policy}{%
\subsection{Updating an existing
policy}\label{updating-an-existing-policy}}

Finally, one can update an existing policy with new observational data.
This is beneficial when, for instance, additional data points become
available or when one wants to plot the reward as a function of the
number of training samples. For this purpose, the
\code{ReinforcementLearning()} function can take an existing \code{"rl"}
model as an additional input parameter. Moreover, it comes with an
additional pre-defined action selection mode, namely
\(\varepsilon\)-greedy, thereby following the best action with
probability \(1 - \varepsilon\) and a random one with \(\varepsilon\).

\begin{Schunk}
\begin{Sinput}
# Sample N = 1000 sequences from the environment
# using epsilon-greedy action selection
data_new <- sampleExperience(N = 1000, 
                             env = env, 
                             states = states, 
                             actions = actions, 
                             actionSelection = "epsilon-greedy",
                             model = model, 
                             control = control)

# Update the existing policy using new training data
model_new <- ReinforcementLearning(data_new, 
                                   s = "State", 
                                   a = "Action", 
                                   r = "Reward", 
                                   s_new = "NextState", 
                                   control = control,
                                   model = model)
\end{Sinput}
\end{Schunk}

The following code snippet shows that the updated policy yields
significantly higher rewards as compared to the previous policy. These
changes can also be visualized in a learning curve via
\code{plot(model\_new)}.

\begin{Schunk}
\begin{Sinput}
# Print result
print(model_new)
\end{Sinput}
\begin{Soutput}
#> State-Action function Q
#>         right         up       down       left
#> s1 -0.6506604 -0.6718149  0.7627013 -0.6704840
#> s2  3.5253380 -0.7694555  0.7197440  0.6965138
#> s3  3.5364934  9.0505592  3.5328707  0.7194558
#> s4 -1.8989173 -1.9085736 -1.9072245 -1.9494559
#> 
#> Policy
#>      s1      s2      s3      s4 
#>  "down" "right"    "up" "right" 
#> 
#> Reward (last iteration)
#> [1] 1464
\end{Soutput}
\begin{Sinput}
# Plot reinforcement learning curve
plot(model_new)
\end{Sinput}

\includegraphics{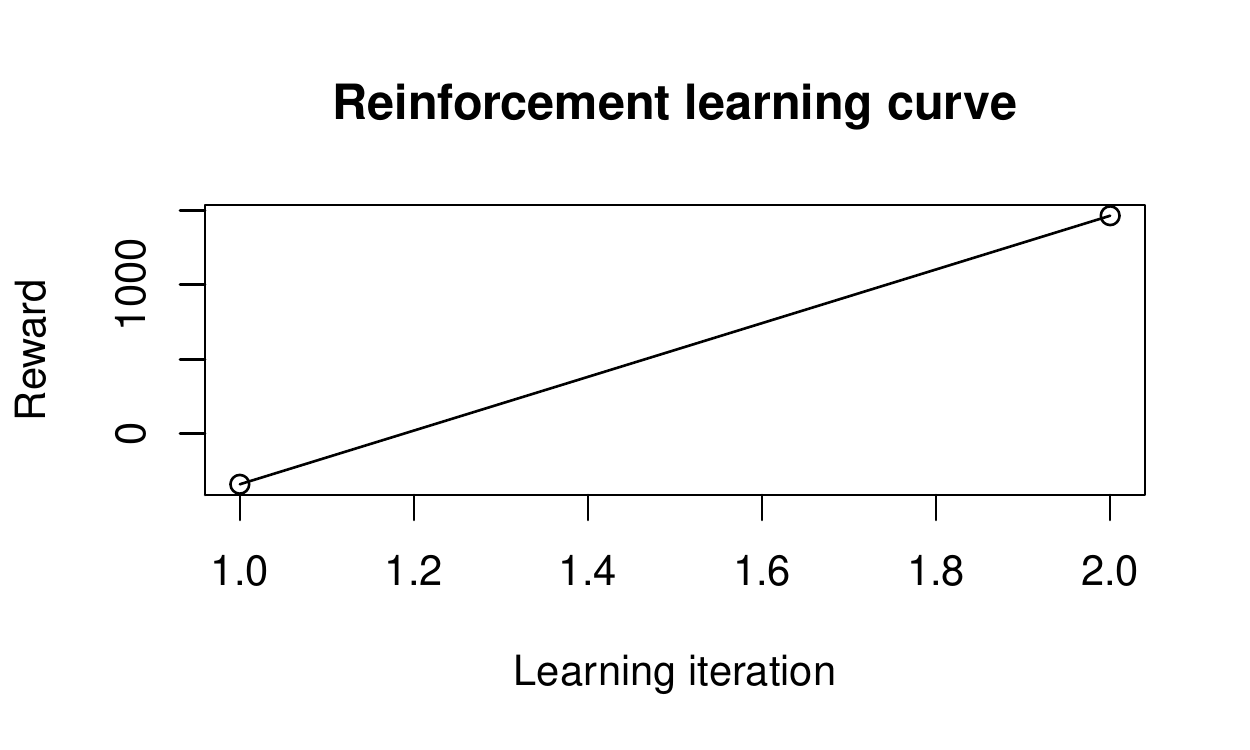} \end{Schunk}

\hypertarget{conclusion}{%
\section{Conclusion}\label{conclusion}}

Reinforcement learning has gained considerable traction as it mines real
experiences with the help of trial-and-error learning. In this sense, it
attempts to imitate the fundamental method used by humans to learn
optimal behavior without the need for of an explicit model of the
environment. In contrast to many other approaches from the domain of
machine learning, reinforcement learning solves sequential
decision-making problems of arbitrary length and can be used to learn
optimal strategies in many applications such as robotics and game
playing.

Implementing reinforcement learning is programmatically challenging,
since it relies on continuous interactions between an agent and its
environment. To remedy this, we introduce the
\pkg{ReinforcementLearning} package for R, which allows an agent to
learn from states, actions and rewards. The result of the learning
process is a policy that defines the best possible action in each state.
The package is highly flexible and incorporates a variety of learning
parameters, as well as substantial performance improvements such as
experience replay.

\bibliography{proellochs-feuerriegel}

\address{%
Nicolas Pröllochs\\
University of Oxford\\
Walton Well Rd, Oxford OX2 6ED, United Kingdom\\
}
\email{nicolas.prollochs@eng.ox.ac.uk}

\address{%
Stefan Feuerriegel\\
ETH Zurich\\
Weinbergstr. 56/58, 8092 Zurich, Switzerland\\
}
\email{sfeuerriegel@ethz.ch}